%% file: conference.tex
\documentclass[conference, anonymous]{IEEEtran}
\IEEEoverridecommandlockouts
\ifCLASSOPTIONcompsoc
\usepackage[nocompress]{cite}
\else
\usepackage{cite}
\fi
\usepackage{amsmath,amssymb,amsfonts}
\usepackage{algorithmic}
\usepackage{graphicx}
\usepackage{textcomp}
\usepackage[utf8]{inputenc}
\usepackage{inconsolata}
\usepackage{tabularx}
\usepackage[T1]{fontenc}
\usepackage{latexsym}
\usepackage{booktabs} 
\usepackage{flushend}
\usepackage{balance}
\usepackage{array} 
\usepackage{float} 
\usepackage{multirow} 
\usepackage{fancyvrb} 
\usepackage[dvipsnames,table]{xcolor}
\usepackage{hyperref} 
\def\BibTeX{{\rm B\kern-.05em{\sc i\kern-.025em b}\kern-.08em
    T\kern-.1667em\lower.7ex\hbox{E}\kern-.125emX}}
\begin{document}

\title{Beyond Ontology in Dialogue State Tracking for Goal-Oriented Chatbot
}


\author{\IEEEauthorblockN{1\textsuperscript{st$^*$} Sejin Lee\textsuperscript} 
\IEEEauthorblockA{\textit{dept. Library and Information Science} \\
\textit{Yonsei Univ.}\\
Seoul, Korea \\
li.sejin@yonsei.ac.kr}
\and
\IEEEauthorblockN{2\textsuperscript{nd$^*$} Dongha Kim\textsuperscript} 
\IEEEauthorblockA{\textit{dept. Library and Information Science} \\
\textit{Yonsei Univ.}\\
\textit{Onoma AI}\\
Seoul, Korea \\
eastha@yonsei.ac.kr}
\and
\IEEEauthorblockN{3\textsuperscript{rd} Min Song}
\IEEEauthorblockA{\textit{dept. Library and Information Science} \\
\textit{Yonsei Univ.}\\
\textit{Onoma AI}\\
Seoul, Korea \\
min.song@yonsei.ac.kr}
}

\maketitle
\def\thefootnote{*}\footnotetext{These authors contributed equally to this work}\def\thefootnote{}
\footnote{This paper was accepted to the 15th IEEE International Conference on Knowledge Graphs (ICKG2024).}

\begin{abstract}
Goal-oriented chatbots are essential for automating user tasks, such as booking flights or making restaurant reservations. A key component of these systems is Dialogue State Tracking (DST), which interprets user intent and maintains the dialogue state. However, existing DST methods often rely on fixed ontologies and manually compiled slot values, limiting their adaptability to open-domain dialogues. We propose a novel approach that leverages instruction tuning and advanced prompt strategies to enhance DST performance, without relying on any predefined ontologies. Our method enables Large Language Model (LLM) to infer dialogue states through carefully designed prompts and includes an anti-hallucination mechanism to ensure accurate tracking in diverse conversation contexts. Additionally, we employ a Variational Graph Auto-Encoder (VGAE) to model and predict subsequent user intent. Our approach achieved state-of-the-art with a JGA of 42.57\% outperforming existing ontology-less DST models, and performed well in open-domain real-world conversations. This work presents a significant advancement in creating more adaptive and accurate goal-oriented chatbots.\footnote{Code and data are available at \url{https://github.com/Eastha0526/Beyond-Ontology-in-DST.git}}
\end{abstract}



\section{Introduction}
Text-based conversational systems, or chatbots, have gained significant attention for their ability to automate user tasks. Goal-oriented chatbots, in particular, specialize in assisting users with tasks such as booking flights or making restaurant reservations \cite{rastogi2020towards,peng-etal-2017-composite}. A key element of these systems is Dialogue State Tracking (DST), which enables the chatbot to maintain an accurate understanding of user goals and dialogue state throughout the conversation \cite{wu-etal-2019-transferable}.

\begin{figure}[t]
  \centering
  \includegraphics[width=\columnwidth, keepaspectratio]{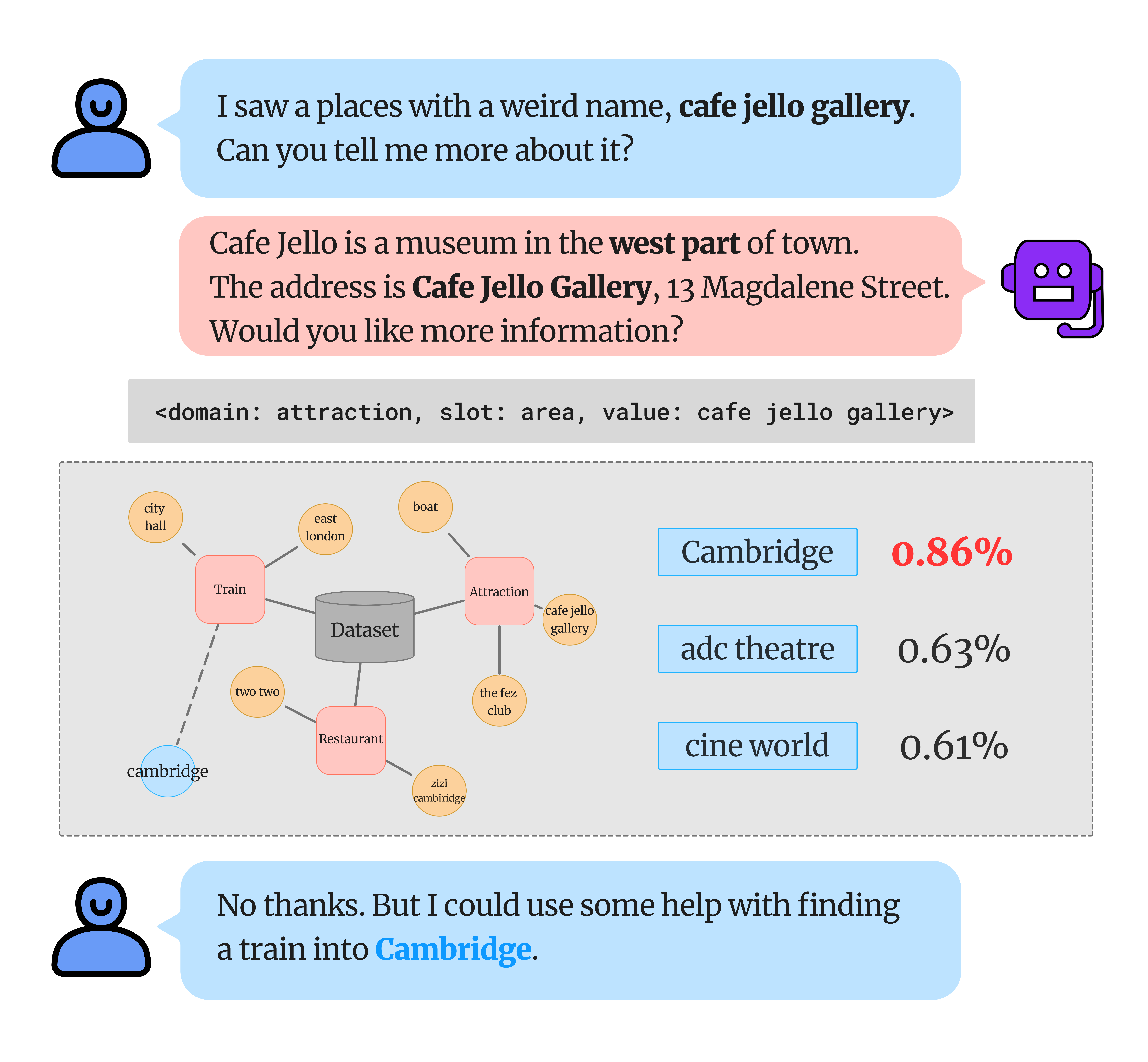}
  \caption{An example of tracking user utterances, representing them as a graph, and predicting the next slot-value. Domains are shown as \textcolor{pink}{rounded squares} and values as \textcolor{orange}{circles}. The model tracks the user utterance `Cafe Jello Gallery' and predicts `Cambridge (\textcolor{cyan}{circle})' as the next slot-value.}
  \label{fig:fig1}
\end{figure}

Traditional DST approaches rely on predefined ontologies—structured, domain-specific vocabularies that define possible intents and slot values. While ontologies provide a framework for organizing dialogue states as triples of \texttt{<domain, slot, value>} (see Figure~\ref{fig:fig1}), they are rigid and resource-intensive, requiring manual updates for each new domain or variation in user input. This makes them unsuitable for open-domain scenarios, where the topics are unpredictable and diverse \cite{gao-etal-2019-dialog,rastogi-etal-2019-scaling}. The process of maintaining an ontology is not only costly but also prone to annotation errors, further complicating its use in real-world applications.

To address these limitations, we propose a novel, ontology-free DST approach that leverages the pre-trained knowledge of Large Language Model (LLM) to dynamically track user goals. LLM-based approaches offer a more flexible solution by leveraging vast pre-trained knowledge, allowing them to handle diverse and dynamic dialogue contexts without relying on predefined schemas. Instead of relying on fixed templates, our method fine-tunes LLM through instruction tuning and prompt design, allowing them to accurately capture dialogue states from the conversation context without predefined slot values. By removing the constraints of manually curated ontologies, our approach provides a more adaptable and scalable solution for real-world conversations. By using techniques such as Chain-of-Thought \cite{wei2022chain}, Persona \cite{white2023prompt}, SELF-DISCOVER \cite{zhou2024self}, and Tree of Thought \cite{yao2024tree}, we create prompts that guide the LLM to extract meaningful dialogue states at each turn. Additionally, we incorporate an anti-hallucination mechanism to prevent the model from producing inappropriate slot values in cases where no valid ontology-free options exist.

\begin{figure*}[t]
  \centering
  \includegraphics[width=\textwidth]{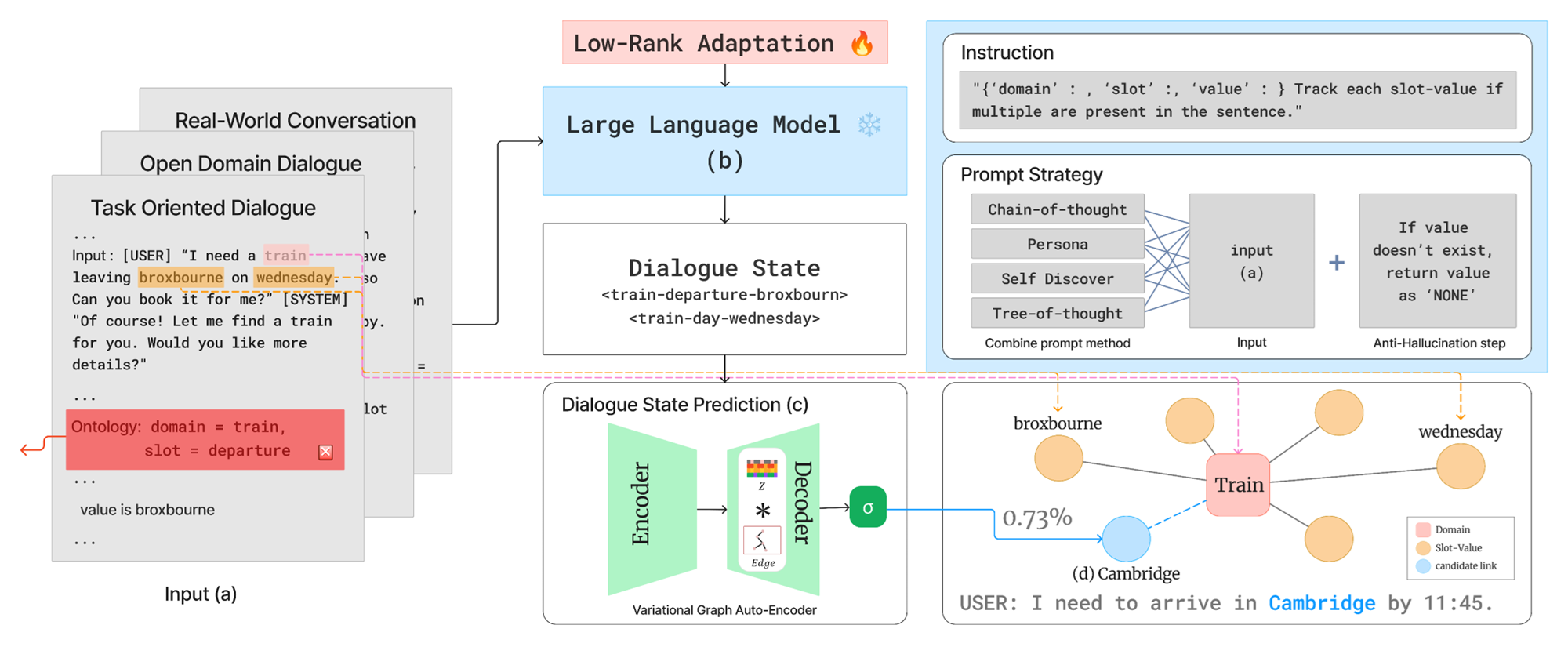}
  \caption{The overall model structure. Given the user dialogue input (a), the instruction and prompt strategy extract the dialogue state with appropriate prompts (b). Here, we design an optimal DST prompt with a prompt strategy based on Chain-of-Thought, input (a), and an anti-hallucination step. Graph the extracted dialogue state into a VGAE (c) to predict the dialogue state (d) that will come from the user's next input.
  }
  \label{fig:fig2}
\end{figure*}

To further enhance dialogue state prediction, we employ a Variational Graph Auto-Encoder (VGAE) to represent dialogue states as graphs. This allows us to reason over subgraphs corresponding to user intents and predict the next dialogue state based on the user's utterance.

Our approach achieves state-of-the-art performance with a JGA of 42.57\%, surpassing existing ontology-less DST models and demonstrating strong performance on open-domain real-world dialogue data(\S \ref{sec:odd}). Our contributions are as follows:
\begin{itemize}
\item We introduce the first ontology-free DST model that accurately tracks dialogue states using advanced prompting techniques, eliminating the need for predefined templates.
\item Our model uses a novel combination of LLM-based prompting and VGAE to predict the next dialogue state in goal-oriented conversations, even in open-domain settings.
\item The proposed method is applicable to unlabeled, real-world conversations, providing a scalable and adaptable solution for DST across diverse domains, thereby contributing to the advancement of goal-oriented chatbots.
\end{itemize}

\section{Background}
In this section, we first introduce the formalism of DST that will be used throughout this paper. We then describe the background of Graph Neural Network (GNN) involved in the process of predicting the next dialogue state, specifically how we use VGAE.
\subsection{Dialogue State Tracking}
DST provides crucial information by tracking the responses exchanged between the user and the system to achieve a specific goal. DST represents the dialogue state as a set of slot-value pairs that reflect the context of the conversation. User utterances are denoted by $U$, and system responses are denoted by $R$. Each turn of the conversation is marked by $t$, and the context up to turn $t$ can be defined as:

$$C_t = \{U_1, R_1, \ldots, R_{t-1}, U_t\}$$

In multi-domain DST scenarios, the user’s intent is represented as domain-slot-value triples, such as \texttt{<restaurant-food-asian>}. The domain M is represented as $\mathcal{D} = \{d_1, d_2, \ldots, d_m\}$, where each domain contains N slots $\mathcal{S} = \{s_1, s_2, \ldots, s_n\}$. The key objective of DST is to accurately identify the slot values $v_{mn}^t$ corresponding to the dialogue context $C_{mn}^t$ for each domain $m$ and slot $n$.
Results from DST can be used as inputs to a GNN to analyze relationships between slot-value pairs within a dialogue state network. Instead of directly enhancing DST, GNN is applied post-DST to model and learn from the structured dialogue data represented in graph form.
For instance, domain and slot values extracted by DST are represented as nodes in a graph $G = (V, E)$, where $V$ represents the domains and slots, and $E$ represents the relationships between them. GNNs utilize this graph structure to predict new relationships (e.g., potential slot-value pairs) or analyze existing ones, allowing for deeper insights into the dialogue state.
A VGAE is well-suited for this context, as it is an unsupervised learning framework capable of capturing hidden patterns in the non-Euclidean space of graph structures derived from DST. It uses latent variables $z$ to encode relationships between nodes (i.e., slot-value pairs), with the goal of predicting new slot-value relationships based on the existing dialogue state graph.
The loss function is defined as:

$$\mathcal{L}_{BCE} = \sum_{i \in V, j \in V} -A_{i,j} \log(\hat{A}_{ij}) - (1 - A_{ij}) \log(1 - \hat{A}_{ij})$$

Additionally, the evidence lower bound (ELBO) loss, incorporating Kullback-Leibler (KL) divergence, is given by:

$$\mathcal{L}_{ELBO} = \mathcal{L}_{BCE} - \mathbf{KL}[q(Z|X,A)$$

Through this process, DST results serve as the foundation for GNN-based models, which can predict new slot-value pairs or infer the user's unspoken intent. By applying GNN, we can analyze the complex structure of dialogue data and gain deeper insights into the results generated by DST.
\input{method}

\section{Experiments setup}
\label{sec:setup}

\subsection{Datasets}
We designed our experiments using three goal-oriented dialogue benchmark datasets: Schema-Guided Dialogue (SGD) \cite{rastogi2020towards}, MultiWOZ2.0 \cite{budzianowski-etal-2018-multiwoz}, and MultiWOZ2.4 \cite{ye-etal-2022-multiwoz}. The MultiWOZ dataset consists of human-to-human conversations and has been revised to correct annotation errors and reduce noise. The SGD dataset comprises approximately 16,000 conversations between humans and a chat assistant, encompassing previously unseen domains and services. 
%
\subsection{Metrics}
We adopt \textbf{slot F1 score} and \textbf{slot accuracy} metrics due to the absence of a predefined template, which makes it challenging to evaluate our model comparably to previous studies that relied on Joint Goal Accuracy (\textbf{JGA}) based on ontology. Nevertheless, we include JGA for consistency with previous DST models. JGA requires all predicted values to exactly match the actual values for accuracy. Slot F1 score measures precision and recall in slot extraction, while slot accuracy assesses the correctness of slot value predictions across conversations. These metrics are suitable for both open-domain and real-world data scenarios.

\subsection{Model}
We used the open-source LLaMA3 model with 8 billion parameters, employing LoRA and instruction tuning. To compare performance differences, we also used GPT-3.5 and GPT-4o with the same instructions and prompts. Training was conducted without the use of ontology, focusing on identifying the best prompt combinations.

\subsection{Graph Construction}
We constructed a graph from the domains and slot-value pairs extracted via DST, linking each domain to its corresponding slot-value pairs. The data was split into 5\% validation, 10\% test, and 85\% training. During training, we masked some graph links to evaluate the model's predictive performance and enhance generalization. Data partitioning was random, and we used 10-fold cross-validation for validation.
\begin{table*}[htbp]
\caption{Evaluation of JGA, slot F1, and slot accuracy for different prompt combinations to find the optimal DST prompt.In all cases, the CoT + persona pattern combination had the best JGA performance across models and datasets. Anti-hallucination phase was included in every case, with the best performance shown in \textbf{bold}.}
\centering
\begingroup
\renewcommand{\arraystretch}{1} 
\resizebox{\textwidth}{!}{%
\begin{tabular}{|l|c|c|c|c|c|c|c|c|c|}
\hline
 &
  \multicolumn{3}{c|}{\textbf{LLaMA3 + Instruction}} &
  \multicolumn{3}{c|}{{\textbf{GPT-3.5}}} &
  \multicolumn{3}{c|}{\textbf{GPT-4o}} \\ \cline{2-10}
 &
  \textbf{\textit{MW2.0}} & \textbf{\textit{MW2.4}} & \textbf{\textit{SGD}} &
  \textbf{\textit{MW2.0}} & \textbf{\textit{MW2.4}} & \textbf{\textit{SGD}} &
  \textbf{\textit{MW2.0}} & \textbf{\textit{MW2.4}} & \textbf{\textit{SGD}}  \\ \hline 
\multicolumn{10}{|c|}{\textbf{Joint Goal Accuracy (JGA)}} \\ \hline
CoT &
  0.4059 & 0.5182 & 0.7100 & 
  0.3931 & 0.4239 & 0.7092 & 
  0.4132 & 0.4837 & 0.7142  \\ \hline
CoT + Persona &
  \textbf{0.4258} & \textbf{0.5664} & \textbf{0.7819} & 
  \textbf{0.4225} & \textbf{0.5247} & \textbf{0.7218} & 
  \textbf{0.4319} & \textbf{0.5345} & \textbf{0.7438}  \\ \hline
SELF-DISCOVER &
  0.3932 & 0.4845 & 0.7539 & 
  0.3325 & 0.5326 & 0.6529 & 
  0.3499 & 0.4943 & 0.6954  \\ \hline
ToT &
  0.2649 & 0.3616 & 0.7690 & 
  0.2974 & 0.3670 & 0.6296 & 
  0.2643 & 0.2834 & 0.6809  \\ \hline
\multicolumn{10}{|c|}{\textbf{Slot F1}} \\ \hline
CoT &
  0.3455 & 0.3868 & 0.7151 & 
  0.3005 & 0.2507 & 0.6766 & 
  0.2636 & 0.3670 & 0.6882  \\ \hline
CoT + Persona &
  \textbf{0.4413} & \textbf{0.4706} & \textbf{0.8002} & 
  \textbf{0.3405} & 0.3411 & 0.7293 & 
  0.3021 & 0.3781 & 0.7293  \\ \hline
SELF-DISCOVER &
  0.3224 & 0.3300 & 0.5976 & 
  0.3236 & \textbf{0.3587} & 0.4407 & 
  \textbf{0.3429} & \textbf{0.3832} & 0.7958  \\ \hline
ToT &
  0.3098 & 0.3154 & 0.7860 & 
  0.2481 & 0.3306 & \textbf{0.7844} &
  0.2149 & 0.2929 & \textbf{0.8223}  \\ \hline
\multicolumn{10}{|c|}{\textbf{Slot Accuracy}} \\ \hline
CoT &
  0.8092 & 0.6837 & 0.5565 & 
  0.8361 & 0.8246 & 0.5113 & 
  \textbf{0.8299} & 0.8097 & 0.5246  \\ \hline
CoT + Persona &
  0.8344 & 0.7902 & \textbf{0.6669} & 
  0.8364 & 0.8334 & 0.5740 & 
  0.8122 & 0.8249 & 0.5740  \\ \hline
SELF-DISCOVER &
  \textbf{0.8757} & \textbf{0.8767} & 0.4261 & 
  0.8223 & 0.8278 &  0.4826 &
  0.8244 & 0.7823 & 0.5531  \\ \hline
ToT &
  0.7971 & 0.7981 & 0.6474 & 
  \textbf{0.9037} & \textbf{0.9214} & \textbf{0.6453}  &
  0.8109 & \textbf{0.9021} & \textbf{0.6983}  \\ 
\hline
\end{tabular}%
}
\endgroup
\label{table:metrics}
\end{table*}
\begin{table*}[ht]
\caption{Comparison of the prediction performance of six GNN models for generating graphs based on extracted dialogue states to predict the dialogue state that will appear in the user's next utterance. Evaluated with 10-fold cross-validation. Our model is VGAE, and the best performance on each metric is shown in bold.}
\centering
\renewcommand{\arraystretch}{1}
\resizebox{\textwidth}{!}{
\begin{tabular}{|l|c|c|c|c|c|c|c|c|c|c|c|c|}
        \hline
             & \multicolumn{2}{c|}{\textbf{GAT}} & \multicolumn{2}{c|}{\textbf{GraphSAGE}} & \multicolumn{2}{c|}{\textbf{GIN}} & \multicolumn{2}{c|}{\textbf{VGAE}} & \multicolumn{2}{c|}{\textbf{CGN-JK}} & \multicolumn{2}{c|}{\textbf{GTN}} \\ \cline{2-13} 
             & \textbf{\textit{AUC}} & \textbf{\textit{AP}} & \textbf{\textit{AUC}} & \textbf{\textit{AP}} & \textbf{\textit{AUC}} & \textbf{\textit{AP}} & \textbf{\textit{AUC}} & \textbf{\textit{AP}} & \textbf{\textit{AUC}} & \textbf{\textit{AP}} & \textbf{\textit{AUC}} & \textbf{\textit{AP}} \\
        \hline
MultiWOZ 2.0 & 0.9324 & 0.9246 & 0.9662 & 0.9634 & 0.9163 & 0.9297 & \textbf{0.9679} & \textbf{0.9798} & 0.7863 & 0.7792 & 0.9378 & 0.9528 \\ \hline
MultiWOZ 2.4 & 0.9307 & 0.9440 & 0.9619 & 0.9661 & 0.9082 & 0.9498 & \textbf{0.9939} & \textbf{0.9954} & 0.9625 & 0.9595 & 0.8561 & 0.9061 \\ \hline
SGD   & 0.9047 & 0.9016 & 0.9859 & 0.9829 & 0.8923 & 0.8743 & \textbf{0.9961} & \textbf{0.9944} & 0.8710 & 0.8756 & 0.9126 & 0.9220 \\ 
\hline
\end{tabular}
}
\label{table:results_auc_ap}
\end{table*}

\begin{table*}[ht]
\centering
\caption{Comparing our approach to other zero-shot DST approaches. Gao and Our approach are the only two DST models without an ontology, and \textbf{Bold} shows the best results.}
\renewcommand{\arraystretch}{0.7}
\resizebox{0.7\textwidth}{!}{
\begin{tabular}{|l|c|c|c|c|}
\hline
           & \textbf{Baseline} & \textbf{MultiWOZ2.0} & \textbf{MultiWOZ2.4} & \textbf{SGD}  \\
\hline
TRADE\cite{wu-etal-2019-transferable} & GRU & 25.76\% & - & -\\ \hline
SUMBT\cite{lee-etal-2019-sumbt} & BERT & 28.18\% & - & -\\ \hline
SimpleTOD\cite{hosseini2020simple} & GPT-2 & 29.65\% & - & - \\ \hline
T5DST\cite{lin2021leveraging} & T5 & 33.56\% & - & - \\ \hline
DiSTRICT\cite{venkateswaran2022district} & T5-small & 38.17\% & - & - \\ \hline
SynthDST\cite{kulkarni2024synthdst} & GPT-3.5 & - & 45.6\% & - \\ \hline
S3DST\cite{das2023s3} & LLM & - & 53.27\% & - \\ \hline
IC-DST\cite{hu-etal-2022-context} & GPT-3.5 & - & 53.60\% & - \\ \hline
ParsingDST\cite{wu-etal-2023-semantic} & GPT-3.5 & - & 64.65\% & - \\ \hline
UNO\cite{li2024unodst}  & T5 & - & - & 47.4\%\\ \hline
TransferQA\cite{lin-etal-2021-zero} & T5 & - & - & 20.7\%\\ \hline
Gao et al., 2019 \cite{gao-etal-2019-dialog} & T5 & 42.12\% & - & -\\ \hline
Ours & LLaMA3-8B & 42.58\% & 56.64\% & 78.19\%\\
\hline
\end{tabular}
}
\label{table:comparison_other_model}
\end{table*}
\begin{table}[ht]
\caption{Comparison of DST performance with zero-shot and few-shot in our approach's Prompt Strategy}
\resizebox{\columnwidth}{!}{
\begin{tabular}{|l|c|c|c|}
\hline
           & \textbf{JGA} & \textbf{Slot F1} & \textbf{Slot Accuracy}\\
\hline
Zero-shot & 0.5664 & 0.3868 & 0.6837 \\ \hline
Two-shot & 0.4527 & 0.3720 & 0.8109 \\ \hline
Three-shot & 0.5333 & 0.3989 & 0.8397 \\ \hline
Four-shot & 0.6175 & 0.3953 & 0.8866 \\ 
\hline
\end{tabular}
}
\label{table:zeroshot}
\end{table}
\section{Result}
We evaluated two separate stages: finding the optimal DST prompts and predicting the next state of the dialogue based on them. We evaluated three models and three datasets with three metrics to find the optimal combination of DST prompts, and the results are shown in Table \ref{table:metrics}. The results of each step are discussed in each section below.
\\
\textbf{Anti-Hallucination Step} Our prompt strategy consisted of a combination of prompt methods, user input, and anti-hallucination steps to improve performance in the absence of an ontology. To evaluate this, we tested our approach using the anti-hallucination step on the MultiWOZ2.4 dataset. The results showed that this technique was effective, with 47.06\% of performances improving the slot F1 score by more than 20\% over the 27.57\% of the prompts without the step. A detailed description of these steps and the results of our experiments can be found in Appendix \ref{appendix:anti}
\\
\textbf{Effective Prompt Strategy}
We hypothesized that our zero-shot inference approach could replace the need for an ontology in an instruction-tuned model, as demonstrated in Table \ref{table:metrics}. Based on the anti-hallucination step, there were four versions of each prompt: CoT, CoT with persona pattern, SELF-DISCOVER, and ToT. Among these, the strategy combining CoT with the persona pattern had the highest JGA value across all conditions. It also achieved the highest slot F1 score, averaging 50.36\% across all experiments. Thus, we found that the CoT with persona pattern was the most effective prompt strategy for the DST task. Variations of the persona pattern can be found in Appendix \ref{sec:appendix prompt}. 
\\
\textbf{Dialogue State Prediction}
%
%
After performing DST with the selected prompts and graphing the tracked dialogue states, we compared different GNNs using AUC and AP metrics to evaluate their performance in predicting the next dialogue state. As a comparison, we selected models that predict links in a graph structure: GAT \cite{velikovi2017graph}, GraphSAGE \cite{hamilton2017inductive}, GIN \cite{xu2018powerful}, CGN-JK \cite{xu2018representation}, GTN \cite{yun2019graph}, and VGAE, the model we used. As shown in Table \ref{table:results_auc_ap}, VGAE performed the best on all datasets, achieving an AUC of 98.6\% and an AP of 98.99\% on average, significantly outperforming other models.

Most GNNs recorded AUC and AP values above 90\%, indicating that the conversation tracking network effectively captured useful link prediction information. These results suggest that our approach is highly effective in graphically organizing dialogue states and using them for advanced link prediction. The exceptional performance of VGAE underscores the importance of graph-based methodologies in tracking and predicting dialogue states.
\\
\textbf{Comparison to the traditional DST model} We compared our approach with the JGA of existing zero-shot DST models evaluated on unseen domains, independent of ontology, as shown in Table \ref{table:comparison_other_model}. The methods compared include TRADE \cite{wu-etal-2019-transferable}, SUMBT \cite{lee-etal-2019-sumbt}, SimpleTOD \cite{hosseini2020simple}, T5DST \cite{lin2021leveraging}, DiSTRICT \cite{venkateswaran2022district}, SynthDST \cite{kulkarni2024synthdst}, S3DST \cite{das2023s3}, IC-DST \cite{hu-etal-2022-context}, ParsingDST \cite{wu-etal-2023-semantic}, UNO \cite{li2024unodst}, TransferQA \cite{lin-etal-2021-zero}, and the ontology-free model of \cite{gao-etal-2019-dialog}. Our model achieved a JGA of 42.57\% on MultiWOZ 2.0, 50.0\% on MultiWOZ 2.4, and 75.8\% on SGD, outperforming ontology-based zero-shot models and reaching performance comparable to few-shot models.

The ontology-less model of \cite{gao-etal-2019-dialog} achieved 39.41\% with a single method and 42.12\% with an ensemble method on MultiWOZ2.0. Our ontology-free approach achieved a state-of-the-art JGA of 42.57\% on the same dataset, demonstrating improved performance over Gao's method. This confirms the validity of our approach in achieving DST goals without relying on ontology.

We maintained strong performance on the MultiWOZ2.4 and SGD datasets, with JGA scores of 56.64\% and 70.99\%, respectively, showing that our model can effectively learn complex conversational patterns. These results suggest that our model can perform well across various real-world conversations, providing important insights into the feasibility of ontology-independent models for DST.
\\
\textbf{Few-Shot Evaluation} Goal-oriented dialogues, like all dialogues, are diverse and broad. To evaluate the generalizability of our ontology-free DST approach, we tested it using a few-shot evaluation: 2-shot, 3-shot, and 4-shot methods compared to the original zero-shot setting.

As shown in Table \ref{table:zeroshot}, the ontology-free approach achieved a superior JGA of 56.64\% in the zero-shot setting. Adding more examples improved slot accuracy, but also increased the likelihood of incorrect answers according to the other two metrics. For example, in the zero-shot setting, we achieved 68\% slot accuracy, which improved to 88\% in the few-shot setting, though the difference in slot F1 score was less than 2\%. This suggests that while additional examples help the model recognize complex patterns, they can also increase the frequency of uncertain answers. Nonetheless, the ontology-free approach appears well-suited for various real-world conversational scenarios.
\\
\textbf{Error analysis} Our LLM-based approach generates more diverse responses than traditional language models but encounters two primary error types: tracking non-existent values and synonym tracking. In the MultiWOZ2.4 dataset, tracking non-existent values accounted for 25.5\% of errors,  such as tracking `...', `XXXXX', and `general'. This highlights challenges in tracking exact values in specific contexts. Synonym tracking errors accounted for 5.3\% of errors, such as tracking `5 nights' instead of `5'. These results illustrate limitations stemming from the absence of ontology and suggest that such errors are inherent to ontology-free models.

\section{Open Domain Dialogue}
\label{sec:odd}
Unlike traditional task-oriented bots, open-domain dialogue systems aim to build long-term relationships with users by fulfilling human needs for communication, affection, and social belonging \cite{huang_etal_2020}. In the previous section, we showed that our approach is completely independent of ontology and does not require labels on the data, and thus can extract dialogue states from open domain dialogue systems and real-world data. To demonstrate this, we used Persona-Chat \cite{zhang-etal-2018-personalizing}, which consists of crowd-sourced conversations in which each participant plays the role of an assigned persona.
\\
We implemented the experiment in the same way as in Section \ref{sec:setup}: we performed the evaluation with completely new data and could only measure slot accuracy, as the slot for the next utterance did not exist. As it turned out, the domain comprising the dialogue was “General”, and predicting the slot for the next utterance based on this yielded an AUC of 0.9589 and an AP of 0.9791. In the absence of metrics, an example of our own human evaluation on a randomized dialogue can be found in Appendix \ref{appendix:odd}. These results imply that our approach is applicable to all common conversations.
\section{Related Work}
\noindent
\textbf{Evolution of Text-Based Conversational Systems} Text-based conversational systems, or chatbots, have become increasingly popular due to their ability to automate user tasks and provide accurate answers. These systems process and understand natural language, track conversation context through a conversation manager, and generate appropriate responses \cite{SUHAILI2021115461}.

To understand natural language and predict dialogue states, DST uses deep neural Network. \cite{wu-etal-2019-transferable} proposed a dialogue state generator using a copying mechanism to address dependency on domain ontology and lack of cross-domain knowledge sharing. \cite{lee-etal-2019-sumbt} introduced the Slot-utterance Matching Belief Tracker (SUMBT) with an Attention mechanism, and \cite{zhu-etal-2020-efficient} used attention mechanisms to efficiently estimate conversational context.
\\
\textbf{Advances in Generation-Based Chatbots and DST} Generation-based chatbots operate across various domains, with research focused on reducing dependence on ontology. \cite{mrksic-etal-2017-neural} introduced a Neural Belief Tracking (NBT) framework using CNNs, and \cite{zhang-etal-2020-find} applied a BERT-style model for non-categorical methods. However, some open-domain chatbots still rely on complex rule-based systems \cite{adiwardana2020towards}.

Zero-shot methodology has been applied to track multi-domain utterances in DST. \cite{aksu-etal-2023-prompter} proposed parameter-efficient transfer learning for zero-shot domain adaptation, and \cite{das2023s3} introduced S3-DST for long context tracking.

With LLM capable of analyzing natural language semantics, studies have applied LLM to DST tasks \cite{feng-etal-2023-towards,king-flanigan-2023-diverse}. Prompting methods include few-shot learning examples \cite{hu-etal-2022-context,madotto2021few}, schema-based prompts \cite{lee-etal-2021-dialogue, zhang-etal-2023-sgp}, instructional prompts \cite{chung-etal-2023-instructtods}, and function calling \cite{li2024large}. 
\\
\textbf{Link Prediction and GNN} Link prediction is essential in network structure data and can be approached with heuristic methods and latent feature extraction \cite{zhang2018link}. GNN are deep learning models that handle graph-related tasks by aggregating information from graph structures and encoding node features \cite{velikovi2017graph}. GNN are effective for link prediction and have been applied to various problems, including DST.

For instance, \cite{lin-etal-2021-knowledge} used Graph Attention Network (GAT) to augment GPT-2 for information extraction from user utterances, and \cite{feng-etal-2022-dynamic} proposed a dynamic schema graph fusion network. \cite{chen2020schema} developed a multi-domain dialogue state tracker using GAT-based schema graphs, and \cite{Zhou2019MultidomainDS} proposed Dialogue State Tracking with Question Answering using bidirectional attention flow.

\section{Conclusion}
We propose a novel approach to extracting dialogue states and predicting the next dialogue state using prompts that replace ontology for goal-oriented conversation agents. We performed well on DST without ontology by tailoring prompts for DST through instruction tuning, prompt strategy. Dialogue state is represented as a relation graph to predict the dialogue state in the user's next utterance to capture the conversation context and user intent. Our experimental results on MultiWOZ and SGD show that our approach performs well without ontology, and the predicted candidate dialogue states allow us to identify the context and user intent in the next conversation. We also verified that our approach performs well on a open domain dialogue dataset called Persona-Chat. We conclude that our approach can be effectively applied to complex dialogue situations in the real world to advance purposeful goal-oriented chatbots.

\section{Limitation}
This study has two key limitations: (1) Our approach relies on zero-shot prompt engineering for DST without using an ontology, combining existing prompting techniques rather than introducing a new algorithm. While the results demonstrate that robust prompts can replace ontologies, future research should focus on developing novel prompting algorithms specifically for DST. (2) We applied our method to open-domain dialogue (\S \ref{sec:odd}), but the absence of a standardized evaluation scale required us to rely on provisional self-validation. To address this, future work will include large-scale human evaluations or the creation of a comprehensive evaluation framework using LLM. These limitations provide important directions for future research and lay the groundwork for developing more advanced and practical DST systems.

\section*{Acknowledgment}
This work was supported by the National Research Foundation of Korea(NRF) grant funded by the Korea government(MSIT) (No. 2022R1A2B5B02002359). This work was partly supported by an IITP grant funded by the Korean Government (MSIT) (No. RS-2020-II201361, Artificial Intelligence Graduate School Program (Yonsei University))

\bibliographystyle{IEEEtran}
\bibliography{IEEEabrv, custom}

\begin{appendices}

\section{Description of Prompt Templates}
\label{sec:appendix prompt}
\subsection{Instruction Tuning}
We used a fixed instruction template to fine-tune an open-source model to create a model for DST. It is important to note that this fixed instruction template does not contain an ontology. This fixed prompt is used in fine tuning, along with the prompts in prompt strategy, and also utilized in the reasoning step. See the example prompt in appendix \ref{appendix:promptstrategy} for an instruction.

\subsection{Prompt Strategy}
\label{appendix:promptstrategy}
Given that traditional DST performs well based on ontology, we aimed to perform DST without using ontology at all. We adopted LLM for its high performance and prompt engineering for its task-specific performance. In the prompt strategy stage, we tried to find the best combination of prompts for the DST task.
We used several prompt engineering methods. The tables below show the combinations using the methods we used. First, we performed the DST with a single prompt to see if each prompt method could work well in the DST. We then performed experiments with multiple variations of each method, including an anti-hallucination step.
We provide some examples of prompts alone and in combination in these prompting strategies below.
\\
\textbf{Prompt Type 1: "CoT"}
\\\{\texttt{\textbf{‘‘prompt’’}:Below is an instruction that describes a task, paired with an input that provides further context. Write a response that appropriately completes the request. Ensure that the response is clear, concise, and directly addresses the task described in the instruction. Avoid asking for personal information or making assumptions beyond the provided context. Let's step by step. \textbf{Instruction}: {instruction} \textbf{Input}: {input} Response:}\}
\\
\textbf{Prompt Type 2: "CoT" + "Persona 1"}
\\\{\texttt{\textbf{‘‘prompt’’}:You are an advanced dialogue state tracker with expertise in understanding and managing complex conversations to maintain context and provide accurate responses. Below is an instruction that describes a task, paired with an input that provides further context. Write a response that appropriately completes the request. Ensure that the response is clear, concise, and directly addresses the task described in the instruction. Avoid asking for personal information or making assumptions beyond the provided context. Let's step by step. \textbf{Instruction}: {instruction} \textbf{Input}: {input} Response:}\}
\\
\textbf{Prompt Type 3: "CoT" + "Persona 2"}
\\\{\texttt{\textbf{‘‘prompt’’}:You are a context-aware dialogue specialist, skilled in recognizing user intents and maintaining seamless conversation flow by accurately tracking dialogue states. Below is an instruction that describes a task, paired with an input that provides further context. Write a response that appropriately completes the request. Ensure that the response is clear, concise, and directly addresses the task described in the instruction. Avoid asking for personal information or making assumptions beyond the provided context. Let's step by step. \textbf{Instruction}: {instruction} \textbf{Input}: {input} Response:}\}
\\
\textbf{Prompt Type 4: "CoT" + "Persona 3"}
\\\{\texttt{\textbf{‘‘prompt’’}:You are an expert conversational analyst, proficient in monitoring and updating dialogue states to ensure coherent and contextually appropriate interactions. Below is an instruction that describes a task, paired with an input that provides further context. Write a response that appropriately completes the request. Ensure that the response is clear, concise, and directly addresses the task described in the instruction. Avoid asking for personal information or making assumptions beyond the provided context. Let's step by step. \textbf{Instruction}: {instruction} \textbf{Input}: {input} Response:}\}
\subsection{Prompt Strategy}
\label{appendix:promptstrategy}
While traditional DST typically performs well based on predefined ontology, our objective was to explore DST without relying on any ontology. To achieve this, we utilized LLM due to its high performance and incorporated prompt engineering techniques to optimize task-specific outcomes. During the prompt strategy phase, we sought to identify the most effective combination of prompts for the DST task.

We employed various prompt engineering approaches and conducted experiments with different combinations. Initially, we performed DST using a single prompt to assess the feasibility of each method. We then tested multiple variations, including the implementation of an anti-hallucination step.

For detailed examples and the full list of prompts used in this study, please refer to our GitHub repository: \href{https://github.com/Eastha0526/Beyond-Ontology-in-DST.git}{https://github.com/Eastha0526/Beyond-Ontology-in-DST.git}.
\section{Additional Result}
\label{sec:appendix result}
\subsection{Anti-Hallucination}
\label{appendix:anti}
%
%
To mitigate hallucinations in LLM, which often generate overly descriptive responses without a defined ontology, we implemented a specific anti-hallucination prompt. For instance, when asked about a "nearby attraction," the LLM might respond with unnecessary details such as "a restaurant called Zizi Cambridge, and a nearby attraction is Cambridge" rather than the concise "Cambridge."

To address this, we introduced a prompt instructing the LLM to return "None" if a precise answer is unavailable. This reduced unnecessary elaboration, yielding more direct responses. The specific prompt was: "If the value does not exist, return the value as NONE." This method significantly enhanced model performance, improving the F1 score by up to 20\%.


Table \ref{table:None with comparsion} show that the Anti-Hallucination step improved slot F1 scores and slot accuracy for different prompt types. The CoT approach, which included the Anti-Hallucination step, was particularly effective. Among the various personas tested, the "Context-Aware Dialogue Specialist" achieved the highest F1 score (47.06\%), making it the best-performing prompt type.

\begin{table}[H]
\caption{\label{table:None with comparsion} Performance comparison of different DST prompt types. The anti-hallucination step was applied to all methods except CoT(Top). Slot F1 and Slot Accuracy are reported for each prompt method, with best and worst performances highlighted.}
\centering
\resizebox{\linewidth}{!}{%
\begin{tabular}{|l|c|c|c|}
\hline
prompt method                                                & \multicolumn{1}{l|}{\textbf{Template}} & \multicolumn{1}{c|}{\textbf{Slot F1}} & \multicolumn{1}{c|}{\textbf{Slot Accuracy}} \\ \hline
CoT(without Anti-Hallucination step) & 1 & \textbf{0.2757} & 0.7806 \\ \hline
CoT & 2 & 0.3547 & \textbf{0.7739} \\ \hline
CoT + Detailed Instruction & 3 & 0.3658 & 0.7809 \\ \hline
Advanced Dialogue State Tracker persona & 1 & 0.3595 & 0.7909 \\ \hline
Context-Aware Dialogue Specialist persona & \textbf{2} & \textbf{0.4706} & 0.7902 \\  \hline
Expert Conversational Analyst persona & 3 & 0.4125 & \textbf{0.8044} \\ \hline
\end{tabular}
}
\end{table}

\section{Open Domain Dialogue}
\label{appendix:odd}
Our approach eliminates the need for ontology or labeled data, making it suitable for open-domain conversations, such as those in chatbots. We validated this using a real open-domain conversation dataset, following the experimental setup described in Section \ref{sec:setup}. As noted in Section \ref{sec:odd}, the dialogues primarily belong to general domains. In total, 4,863 slots and 20,315 values were extracted through DST, with approximately 58.62\% of the values classified as 'NONE'. Our model predicted dialogue states based on general domains, achieving an AUC of 0.9589 and an AP of 0.9791, demonstrating the robustness of the approach.

Evaluating DST in open-domain dialogues poses unique challenges, as these dialogues lack predefined ontologies and standardized evaluation metrics. To address this, we propose human evaluations of the results generated by our model. Table \ref{table:odd_dst_example} presents a comparison of extracted DST, while Table \ref{table:conversation_example} contrasts the generated candidate conversation states with the corresponding actual utterances. The highest-ranked candidate link is highlighted in \textbf{bold}, with the subsequent conversation in \textit{italics}.

\fontsize{10pt}{12pt}\selectfont
\begin{table}[H]
  \caption{\label{table:odd_dst_example} Example of applying DST in open domain dialogue: Given a conversation, our ontology-free approach tracks domain, slot, and value.}
  \centering
  \renewcommand{\arraystretch}{1}
  \begin{tabularx}{\linewidth}{|>{\raggedright\arraybackslash}X|}
    \hline
    `I am ! for my hobby I like to do canning or some whittling .'\newline
    \textbf{Domain : [`General'] , Slot : [`hobby'] , Value : [`canning or whittling']}  \newline
    `I would have to say its prime rib . Do you have any favorite foods ?' \newline
    \textbf{Domain : [`Restaurant'] Slot : [`food'] Value : [`prime rib']}
    \\ \hline
    `Do you have anything planned for today ? I think I am going to do some canning .'\newline
    \textbf{Domain : [`General'] Slot : [`plan', `activity'] Value : [`canning', `NONE']}\newline
    `Wow , four sisters . just watching game of thrones .' \newline
    \textbf{Domain : [`General'] Slot : [`topic'] Value : [`Game of Thrones']}
    \\ \hline
  \end{tabularx}
\end{table}
\begin{table}[H]
  \caption{\label{table:conversation_example} Examples of applying DST in open domain dialogue using Persona Chat to infer the interlocutor's Goal. \textbf{Bold} text indicates correctly inferred goal.}
  \centering
  \renewcommand{\arraystretch}{1}
  \begin{tabularx}{\linewidth}{|>{\raggedright\arraybackslash}X|}
    \hline
    [USER] `Sometimes, dogs scare me if I do not have a relationship with them .' \newline
    [SYS] `I think that is why I am single, I have 5 dogs .' \newline
    \textbf{Node pair: ([`General'], ['FEAR-DOGS']), Probability: 0.7311} \newline
    Node pair: ([`General'], [`NONE-NONE']), Probability: 0.7310 \newline
    Node pair: ([`General'], [`Financial-Employment']), Probability: 0.7309 \newline
    \textit{`Maybe you should hang out in a \textbf{dog} park and find another \textbf{dog} person .'}\\
    \hline
    [USER] `Oh what did you cook ?'\newline
    [SYS] `I made chicken parmesan and baked a pie .'\newline
    \textbf{Node pair: ([`General'], [`food-apple pie']), Probability: 0.7309} \newline
    Node pair: ([`General'], [`food-bat meat']), Probability: 0.7308 \newline
    Node pair: ([`General'], [`food\_order-medium well']), Probability: 0.7308 \newline
    \textit{`That sounds really tasty ! Was it an \textbf{apple pie} ?'}\\ \hline
  \end{tabularx}
\end{table}
\end{appendices}
\end{document}

%% file: method.tex
\section{Methodology}
Our overall process is provided in Figure  ~\ref{fig:fig2}. Taking unstructured current dialogue as input, the LLM infer the optimal DST prompt through instruction and prompt strategy. At this stage, prompt design becomes an excellent strategy to replace ontology. The extracted dialogue state is structured into a graph through the VGAE stage, ultimately predicting the dialogue state to be extracted from the user's next utterance.
\subsection{Instruction Tuning}
To improve LLM’s DST reasoning without relying on an ontology, we performed instruction tuning. Our approach provides the model with explicit and precise instructions, enabling it to accurately extract domains, slots, and values across various contexts. The model is trained to follow the given instructions, allowing it to consistently track the dialogue state in a conversation scenario without the need for an ontology. For more details, see Appendix \ref{sec:appendix prompt}.

We also utilized one of the parameter-efficient fine-tuning methods, the Low-Rank Adaptation (LoRA) technique \cite{hu2021lora}. This allowed us to improve the model's learning speed and achieve high performance with fewer resources. It enables the model to properly extract dialogue states in different conversational situations and accurately understand user intent in multi-domain dialogues.
\subsection{Prompt Strategy}
Previous work has incorporated ontology to constrain output, but our approach eliminates the need for ontology entirely. Our objective is to fully replace the role of ontology, which has traditionally enhanced DST performance. To achieve this, we integrate a thought-based method aimed at discerning latent user intentions during dialogue. We construct our prompting templates using Chain-of-Thought (CoT) reasoning, which guides the LLM through a series of intermediate reasoning steps, improving its ability to handle complex tasks and better understand user intentions \cite{wei2022chain}.

Several prompting methods have evolved from CoT, including Tree of Thoughts (ToT) \cite{yao2024tree} and SELF-DISCOVER \cite{zhou2024self}. ToT generalizes CoT by enabling the exploration of coherent text units, while SELF-DISCOVER allows the LLM to autonomously uncover intrinsic reasoning structures within tasks, improving performance in domain-specific applications. Additionally, \cite{white2023prompt} demonstrated that assigning a persona to an LLM enhances its task performance. We incorporate a persona into our prompts, improving the model's ability to track the dialogue state accurately.

Our focus is on optimizing prompt formats for the DST task without relying on ontology. Previous research \cite{wei2021finetuned} has shown that zero-shot learning can outperform few-shot learning in fine-tuned language models. Thus, we adopt zero-shot DST, using a fine-tuned language model applicable to open-domain dialogue systems and real-world chatbots without the need for a predefined dictionary. By consolidating these techniques, we design prompts that effectively substitute for ontology-based conceptualization and relationships between concepts.
\\
\\
\textbf{Anti-Hallucination}
Because the ontology-free approach does not require labeled data and allows LLM to track unstated values, it can significantly increase the risk of hallucinations. The anti-hallucination step enhances performance by incorporating the possibility of a correct answer in the prompt, rather than relying on fine-tuning \cite{slobodkin-etal-2023-curious}. We adopted this step to prevent the tracking of non-existent values and to ensure that inferences are appropriate for the DST task. Hallucinations can be reduced by including a brief, explicit instruction: \texttt{``If the value does not exist, return the value as NONE.''}

This anti-hallucination technique not only enhances the LLM’s performance in DST by reducing its reliance on ontology but also ensures that our ontology-free approach remains effective across unseen domains, such as open-domain dialogue and real-world data.

%
%
\subsection{VGAE Stage}
We propose a VGAE stage that constructs a graph structure from the set of dialogue states extracted in the previous stage to predict the goal of the next utterance. Domains and slot-value pairs are represented as nodes in the graph, with edges indicating their relationships. The slot-value nodes represent the user’s semantic intent. By utilizing VGAE, we can generate candidates for the next utterance's goal based on the previously extracted dialogue states. These candidate links predict multiple potential purposes in the user's subsequent utterances.

To predict the user’s next dialogue goal, a subgraph is constructed by extracting dialogue states from the given dialogue. The encoder generates a latent representation, $\hat{z}$, from the subgraph. The decoder then predicts links between nodes using the inner product of $E$ and $\hat{z}$. The resulting probabilities, derived via a sigmoid function, identify the most likely candidate dialogue states for the user’s subsequent utterances.